# Automated diagnosis of pneumothorax using an ensemble of convolutional neural networks with multi-sized chest radiography images

Tae Joon Jun, Dohyeun Kim, and Daeyoung Kim

*School of Computing, KAIST, Daejeon, Republic of Korea*

Daeyoung Kim, kimd@kaist.ac.kr, Professor of School of Computing, KAIST, Daejeon, Republic of Korea

# Automated diagnosis of pneumothorax using an ensemble of convolutional neural networks with multi-sized chest radiography images

Pneumothorax is a relatively common disease, but in some cases, it may be difficult to find with chest radiography. In this paper, we propose a novel method of detecting pneumothorax in chest radiography. We propose an ensemble model of identical convolutional neural networks (CNN) with three different sizes of radiography images. Conventional methods may not properly characterize lost features while resizing large size images into 256 x 256 or 224 x 224 sizes. Our model is evaluated with ChestX-ray dataset which contains over 100,000 chest radiography images. As a result of the experiment, the proposed model showed AUC 0.911, which is the state of the art result in pneumothorax detection. Our method is expected to be effective when applying CNN to large size medical images.

Keywords: pneumothorax; chest radiography; convolutional neural network

**Introduction**

Pneumothorax is caused by the presence of air between the parietal and visceral pleura. This is a relatively common respiratory disease that can occur in a wide range of patients and in various clinical settings. Symptoms of pneumothoraces range from mild pleurisy discomfort and breathing to severe cardiovascular disruption and life-threatening cases require immediate medical intervention and prevention (Currie et al. 2007). Diagnosing a pneumothorax in a chest radiography image is not difficult for an experienced physician, but in some cases, it can easily be missed. Especially, radiography image-level pneumothorax diagnosis is meaningful as a pre-stage for computed tomography (CT) imaging for more clear pneumothorax confirmation. In other words, a machine learning-based pneumothorax diagnosis technique from the chest radiography image is required to assist a physician to diagnose a pneumothorax.

Several methods have been presented in the literature for classifying pneumothorax from the ChestX-ray8 dataset. Wang has released a ChestX-ray8 dataset with eight labeled thoracic diseases in (Wang et al. 2017). The ChestX-ray8 dataset contains more than 100,000 chest radiography images that are multi-labeled with atelectasis, cardiomegaly, effusion, infiltration, mass, nodule, pneumonia, and pneumothorax. In this study, Wang measured classification performance on pre-trained AlexNet (Krizhevsky, Sutskever, and Hinton. 2012), GoogLeNet (Szegedy et al. 2015), VGGNet16 (Simonyan and Zisserman. 2014), and ResNet (He et al. 2016), but the AUC of diagnosing the pneumothorax was only 0.806 due to lack of sophisticated optimization. The limitations of Wang's optimization method are described in more detail in the Background section. Yao proposed a two-stage neural network model that combines a densely connected image encoder with a recurrent neural network decoder to classify thorax diseases from the ChestX-ray8 dataset in (Yao et al. 2017). Yao obtained improved classification performance compared to Wang's study and the AUC of detecting pneumothorax was 0.841. Most recently, Rajpurkar applied a 121-layer deep structure DenseNet (Huang et al. 2017) to the ChestX-ray8 dataset, and the AUC of the pneumothorax detection was 0.888, which was better than previous Wang and Yao's (Rajpurkar et al. 2017). Our proposed method shows a 0.911 AUC in the diagnosis of pneumothorax, which is superior to the previous three studies.

In this paper, we propose a convolutional neural network (CNN) classifier for diagnosing pneumothorax in chest radiography images. We propose an ensemble model of identical CNN models with multi-sized chest radiography images to efficiently classify relatively large chest radiography images. We resized 1024 x 1024 original radiography images into three different sized images of 512 x 512, 384 x 384, and 256 x 256, then use them as input for each of the three identical 50-layer CNN models. The three CNN models determine the final class in the form of an ensemble that averages the output probabilities of each softmax layer. When the image is resized to three different sizes, there is a difference in the matrix size just before the fully-connected layer of the CNN model. In a 512 × 512 sized input, the probability of softmax layer is calculated by training relatively general features of the original radiography image. On the other hand, in a 256 x 256 sized input, the probability is calculated based on relatively specific features of the original image. And in a 384 x 384 image, it is the middle of the two. Therefore, by training CNN models through images of various sizes, it is possible to minimize lost features of the original image and to expect higher performance than using a single sized image as an input. As result, we obtained 0.911 Area Under the ROC Curve (AUC) that is 0.023 higher than the result of learning the CNN model by resizing to a single 256 x 256 size. The 0.911 AUC is the highest performance classifying pneumothorax in the Chest X-ray dataset when compared to previous studies. In addition, our proposed multi-sized images as inputs of CNN models can be effectively applied to medical images with high resolution such as fundus, CT, and Magnetic Resonance Imaging (MRI) images.

The rest of this paper is organized as follows. Section 2 provides background information on the convolutional neural network and pneumothorax. Section 3 explains the detailed methodologies used for the pneumothorax detection. Evaluation and experimental results are in Section 4. Finally, Section 5 draws the conclusion and future work of the paper.

**Background**

*Convolutional neural networks*

CNN applies convolution filter, commonly used in image processing and signal processing, to deep learning, thereby lowering the complexity of the model and extracting fine characteristics of the input data. In particular, CNN exhibits excellent classification performance in the image classification field because the convolution layer extracts characteristics of neighboring elements of the matrix structure well. Input values passed through the convolution layer are filtered through the pooling layer to eliminate noise and extract summarized results from the previous feature map. Figure 1 shows an example of a 2-D input passing through the 3x3 convolution layer and the 2x2 max pooling layer in CNN.

CNN was first introduced by LeCun et al in 1989 (LeCun et al. 1989) and was developed through a project to recognize handwritten zip codes. There are several successful CNN models of the ImageNet Large Visual Perception Challenge (ILSVRC) (Russakovsky et al. 2015), a competition for detecting and classifying objects in a given set of images. AlexNet (Krizhevsky, Sutskever, and Hinton. 2012) announced in 2012, took first place with overwhelming performance as the first model to use the CNN model and GPU at ILSVRC. In 2014, GoogLeNet (Szegedy et al. 2015) and VGGNet (Simonyan and Zisserman. 2014) are the first and second in the ILSVRC, respectively. Although VGGNet is in second place, the VGGNet structure with repeating structure of 3x3 filters and subsampling is more often used in image recognition because the structure is much simpler and performance is not much different from GoogLeNet. ResNet (He et al. 2016) and DenseNet (Huang et al. 2017), which appeared in recent, were proposed to solve the problem that the effect of the early features of the image on the final output is small when the depth of the CNN model is deeper.

We expected that ResNet and DenseNet would perform similarly if they were well optimized for CNN model training. In fact, Rajpurkar's approach (Rajpurkar et al. 2017) with DenseNet and our single ResNet experiment showed similar AUC. Wang

(Wang et al. 2017) also used the same ImageNet pre-trained 50-layer ResNet, but the lower AUC for pneumothorax diagnosis was because he used transfer-learning method to learn only the top layers. This method is appropriate when you need to learn a small number of images on a deep CNN model, such as an Iris dataset, but not on more than 100,000 images, like a ChestX-ray8 dataset. A relatively large dataset, such as ChestX-ray8, needs to re-train the upper convolution layer of the pre-trained CNN model, and the appropriate layer depth must be known through the model optimization process.

*Pneumothorax*

Pneumothorax can be classified as primary, secondary, iatrogenic or traumatic depending on the aetiology (Currie et al. 2007). Primary spontaneous pneumothorax is most common in young, tall, and thin males, and has no history of thoracic trauma or predisposing lung disease. In many cases, rupture of an underlying small subpleural bleb or bulla is thought to be responsible for primary pneumothorax. The occurrence of secondary pneumothorax is caused by underlying lung abnormalities. Pathogenesis of secondary pneumothorax includes chronic obstructive pulmonary disease, bronchiectasis, lung cancer, pneumonia, and pulmonary fibrosis (Currie et al. 2007). An iatrogenic pneumothorax is mainly caused by central vein cannulation, pleural tap or biopsy, transbronchial biopsy, fine needle aspiration, and has occasionally been caused by acupuncture (Currie et al. 2007). Traumatic pneumothorax occurs after direct damage to the thorax. Common causes include penetrating chest injury or a fractured rib lacerating the visceral pleura.

The diagnosis of pneumothorax is usually made by radiography or CT imaging. Posterior-anterior chest radiography images are generally taken to show from the edge of the visceral pleura to the chest wall. As such, the chest radiography image should not be distorted in the image augmentation process, which will be described in the

Methodology section, because the region to be observed is specifically specified. For example, augmentation by rotating the image at random angles may be useful for analyzing cross-sectional images of coronary arteries, but not for chest radiography image analysis. Figure 2 shows a normal posterior-anterior chest radiography (left) and shows chest radiography with a large right-sided spontaneous pneumothorax (right).

**Methodology**

*Radiography image augmentation*

Data augmentation is the best way to prevent the model from falling into overfitting. Especially when CNN is used as a classifier, it is important to perform augmentation carefully that does not significantly modify the characteristics of the original image. The posterior-anterior chest radiography image used for the detection of pneumothorax determines the presence of pneumothorax according to the contrast. Therefore, the channel shift method, which maintains the difference in contrast, is highly preferred. In addition, the chest radiography image is taken to include lung apex to costophrenic angle (CPA), so flipping the left and right side of the image, rescaling the image size, and cropping the part of the image are can be applied unless they do not deviate much from the essential parts of the chest. For every epoch, the radiography image is augmented according to the given random parameters. The range of the given random parameter is as follows. The rescaled range of the image was set from 0.875 to 1.125, and the cropping range was set to a range not exceeding 0.125 in the width of the original image. Figure 3 shows the augmented images with the channel shift applied and the original image (leftmost).

*Pneumothorax classifier*

The CNN classifier we propose for the diagnosis of pneumothorax is an ensemble of three identical 50-layer ResNets with different image input sizes. The size of chest radiography image we are going to train is 1024 x 1024, which is larger than the image used for general image classification. However, most of the images take a similar shape, and the part that distinguishes between normal and pneumothorax is only part of the image. Therefore, resizing an input image with multiple sizes, such as 512 x 512, 384 x 384, and 256 x 256 can preserve the lost information in the resizing process. Three ResNet models are trained based on each image-sized input, and the final prediction result is calculated with the average of the softmax outputs of the three models. Figure 4 shows the architecture of the proposed CNN classifier for pneumothorax diagnosis.

The weights of the ResNet were initialized from the weights of the pre-trained model on ImageNet. We included an average pooling layer after the output of the pre-trained model, followed by dense layer size of two that is the number of classes (normal and pneumothorax). The network is trained using RMSprop optimizer (Tieleman and Hinton. 2012) with an initial learning rate of 0.0001 that is decayed by a factor of 1e-8 for each epoch. Since Wang fixed the weight of the pre-trained model on ImageNet and only trained layer after global average pooling, the model did not show reasonable classification performance. Of course, if the size of the dataset is small, it is possible to fix the weights of the pre-trained model and learn only the top-layer. However, since the ChestX-ray dataset requires to train near 100,000 images, the method is not suitable. Therefore, we used a method of fine-tuning the two top-dense layers first and fully training the entire network (Yosinski et al. 2014). Other parameters considered for model optimization are as follows. The drop rate, which is the probability of dropping the unit of the dense layer, is set to 0.5 (Dropout 2012). All of the activation functions of the model were Rectified Linear Unit (ReLU) (Nair and Hinton 2010).

**Experiments and Results**

*Experimental setup*

Evaluation of the pneumothorax classification was performed using the ChestX-ray dataset released by Wang (Wang et al. 2017). The ChestX-ray dataset contains over 100,000 chest radiography images from more than 30,000 patients and we used 59,156 normal images and 5,225 pneumothorax images from the dataset. A total of 64,381 chest radiography images were divided into train-set and test-set at 80: 20 ratio. For the early stopping of the model, we set the validation-set to 20% of the train-set and stop the training process if there is no improvement in validation loss during 3 epochs.

The implementation of the CNN model was developed in python language and Keras library (Chollet 2015) using TensorFlow (Abadi et al. 2016) as backend was mainly used. The CNN model was trained on a server with two NVIDIA Titan GPUs supporting CUDA 9.0 (NVIDIA 2018).

*Evaluation metrics*

The evaluation of the classification considered four metrics: Specificity (Sp), Sensitivity (Se), Accuracy (Acc), and AUC. The AUC is an area under the Receiver Operating Characteristics (ROC) curve. The AUC is calculated by Riemann sum with a set of thresholds to compute pairs of True Positive Rate (TPR) and False Positive Rate (FPR). Hosmer and Lemeshow provided the guidelines for rating the AUC values (Hosmer, Lemeshow, and Sturdivant 2013). Table 1 describes the brief guideline rules of the AUC interpretation.

Specificity is the fraction of negative test results that are correctly identified as normal. Sensitivity is the probability of positive test results that are correctly identified as pneumothorax. Accuracy is the average of specificity and sensitivity. Specificity and sensitivity are defined with four measurements in following:

- True Positive (TP): Correctly detected as pneumothorax
- True Negative (TN): Correctly detected as normal
- False Positive (FP): Incorrectly detected as pneumothorax
- False Negative (FN): Incorrectly detected as normal

Using the above measurement, specificity (Sp) and sensitivity (Se) are defined in following:

$$\text{Sp}(\%) = \frac{TN}{TN + FP} \times 100 \quad (1)$$

$$\text{Se}(\%) = \frac{TP}{TP + FN} \times 100 \quad (2)$$

There is a trade-off between specificity and sensitivity. When the softmax layer is used as the final prediction layer, it classifies the input as the class with the highest probability. The problem is that medical data generally consists of multiple normal and few abnormal (disease) data. In this case, training is processed in the direction of reducing the loss of normal which occupies the majority in the mini-batch. Therefore, when the probability of softmax is directly used for class determination, extremely high specificity and low sensitivity are obtained. To avoid this problem, we chose the cut-off value as the point where the sum of specificity and sensitivity in the ROC curve is the maximum.

*Evaluation results*

Table 2 compares the AUC results of pneumothorax classification with previous studies. From the Table 2, we can see that the proposed method has a 0.023 higher AUC than the existing state-of-art result.

Table 3 compares the evaluation results of the proposed method with the results of the individual models without ensemble. From the Table 3, we can observe that the average AUC result of the individual models are about 0.894 AUC that is similar to the existing state-of-art result, and there is an increase of 0.016 AUC through the ensemble. Our cut-off criterion is the point at which the sum of specificity and sensitivity is the maximum, which is consistent with the point at which the accuracy is maximized. Therefore, based on the accuracy, the performance improvement is about 1.3% through the ensemble. Figure 5 shows the ROC curves for each model. From the Figure 5, we can also observe that the ROC curve of the ensemble model exists above the curve of the other models.

**Conclusion**

Radiography image-level pneumothorax diagnosis is meaningful as a pre-stage for CT imaging for more clear pneumothorax confirmation. In other words, a machine learning-based pneumothorax diagnosis technique from the chest radiography image is required to assist a physician to diagnose a pneumothorax. In recent years, studies have been actively conducted to diagnose thoracic disease by using deep learning, especially convolutional neural network, with ChestX-ray dataset. We propose an ensemble model of identical 50-layer ResNet models with three-sized chest radiography images to efficiently classify relatively large chest radiography images. As a result, we showed pneumothorax classification results that exceeded the existing state-of-art results. As part of the future work, we plan to develop a service that will help physicians diagnose pneumothorax with the proposed CNN classifier.

Acknowledgements

This work was supported by Institute for Information & communications Technology Promotion (IITP) grant funded by the Korea government (MSIP) (No.2016-0-00067, Wise-IoT)

Table 1. AUC interpretation guidelines.

| AUC | Guidelines |
|---|---|
| 0.5 – 0.6 | No discrimination |
| 0.6 – 0.7 | Poor discrimination |
| 0.7 – 0.8 | Acceptable discrimination |
| 0.8 – 0.9 | Good discrimination |
| 0.9 – 1.0 | Excellent discrimination |

Table 2. AUC results of pneumothorax classification.

| Pathology | Wang | Yao | Rajpurkar | Proposed |
|---|---|---|---|---|
| Pneumothorax | 0.806 | 0.841 | 0.888 | 0.911 |

Table 3. Detailed evaluation results of pneumothorax classification.

|  | AUC | Sp (%) | Se (%) | Acc (%) |
|---|---|---|---|---|
| Ensemble | 0.911 | 84.02 | 85.51 | 84.77 |
| Model 512 | 0.897 | 84.21 | 82.58 | 83.40 |
| Model 384 | 0.888 | 83.72 | 81.32 | 82.52 |
| Model 256 | 0.898 | 83.66 | 83.07 | 83.37 |

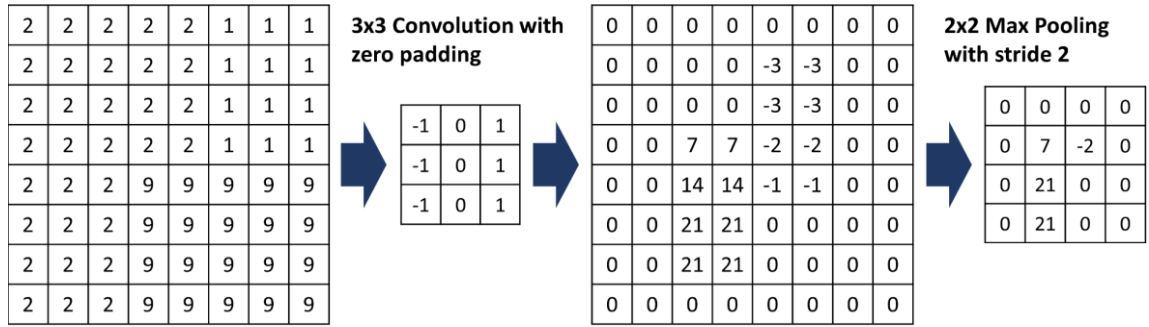

Figure 1. Example of convolution and max pooling layers in CNN.

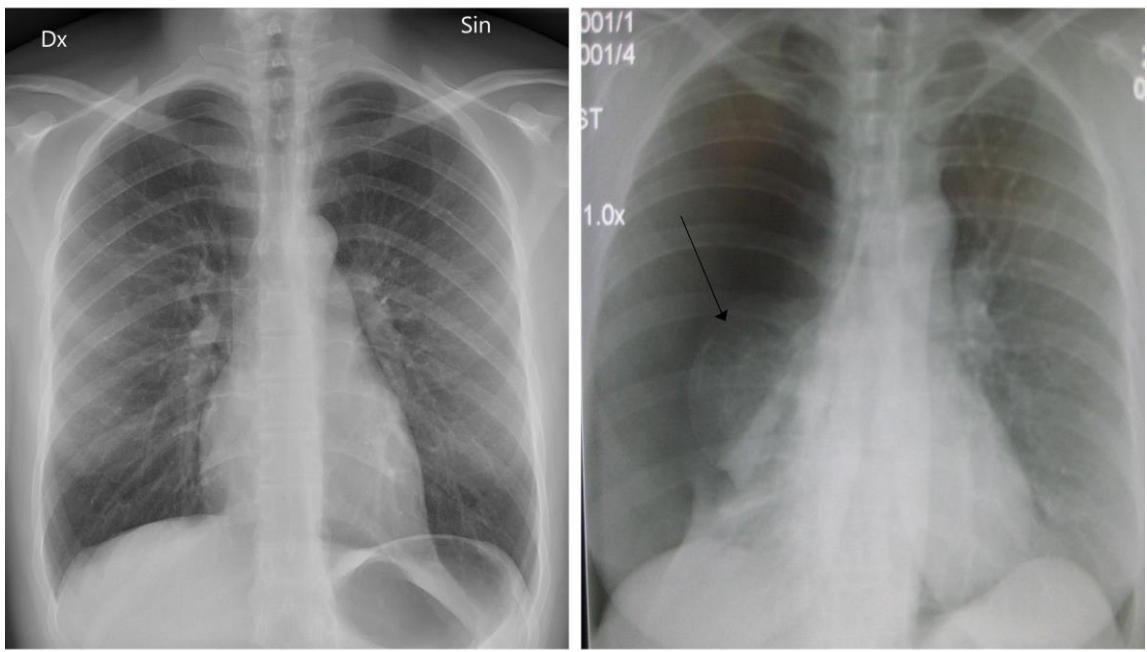

Figure 2. Normal chest radiography (left) and radiography with a pneumothorax (right).

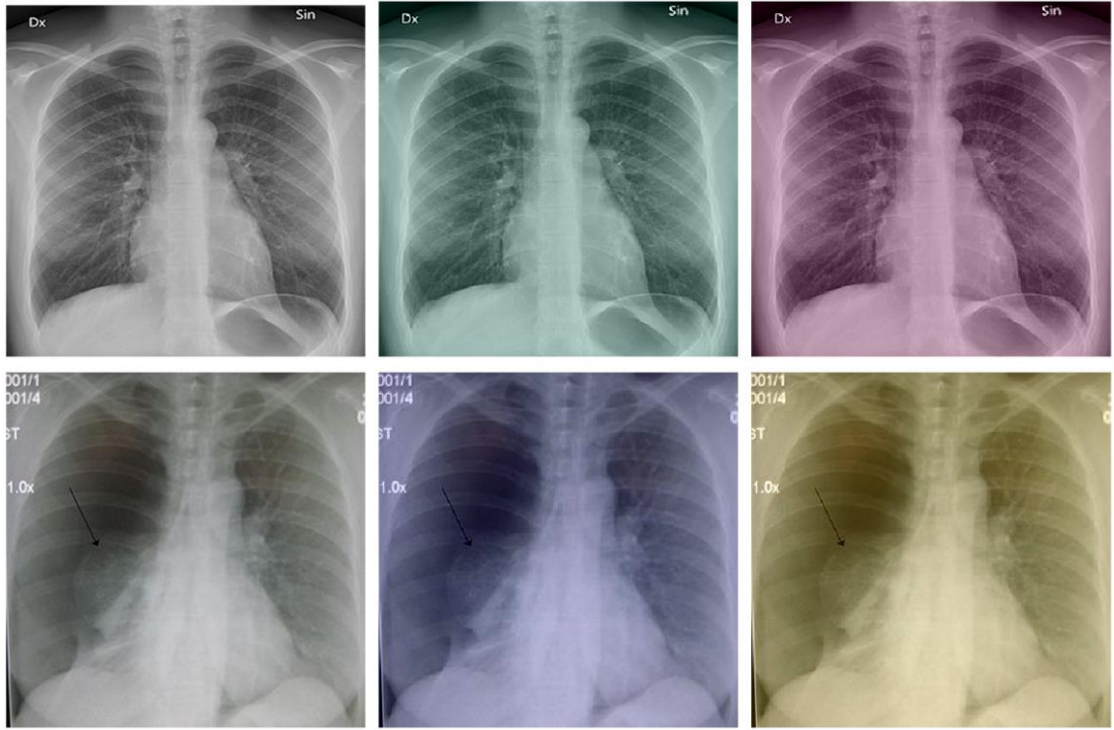

Figure 3. The augmented images with the channel shift applied and the original image (leftmost).

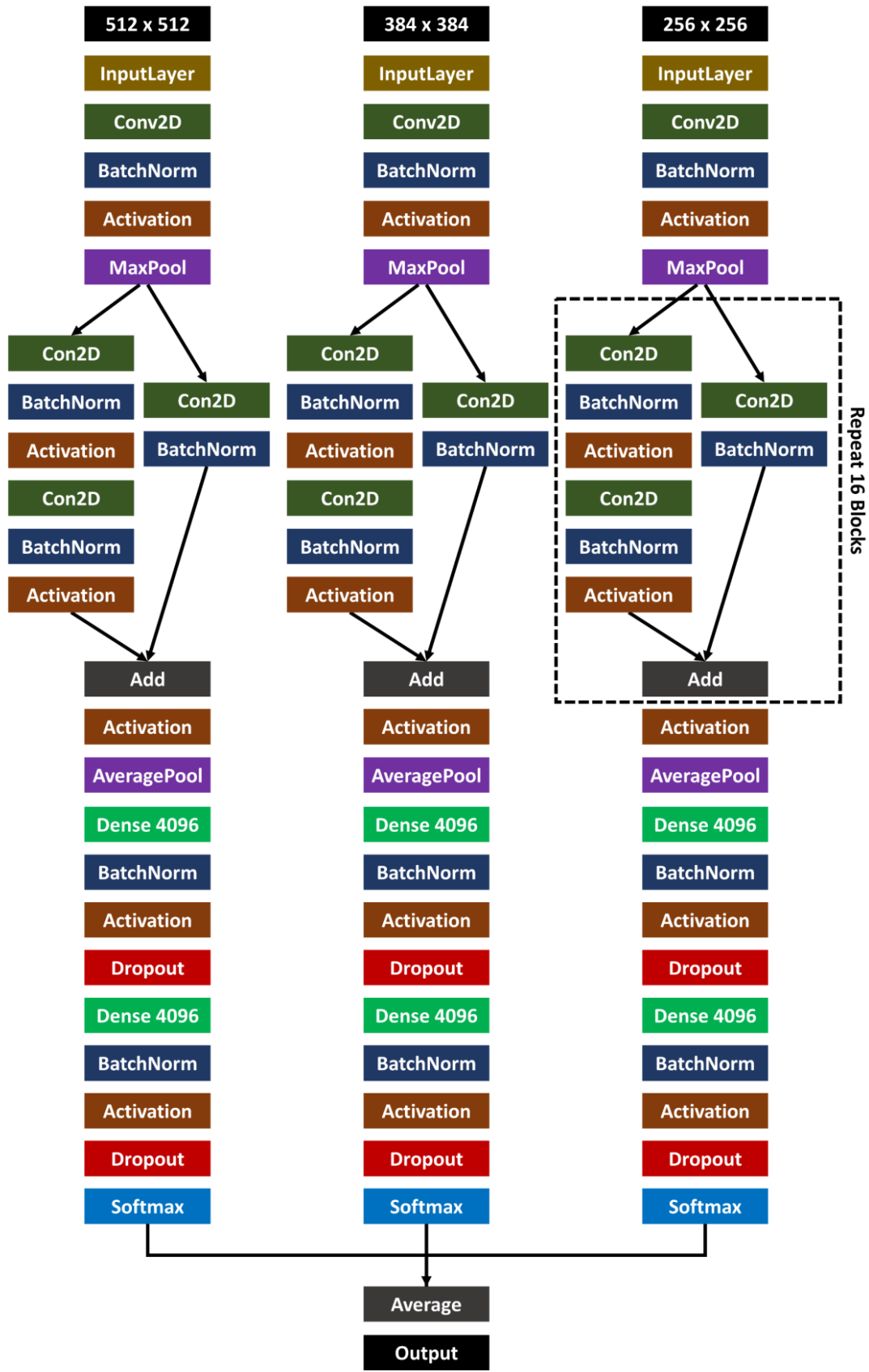

Figure 4. The architecture of the proposed CNN classifier for pneumothorax diagnosis.

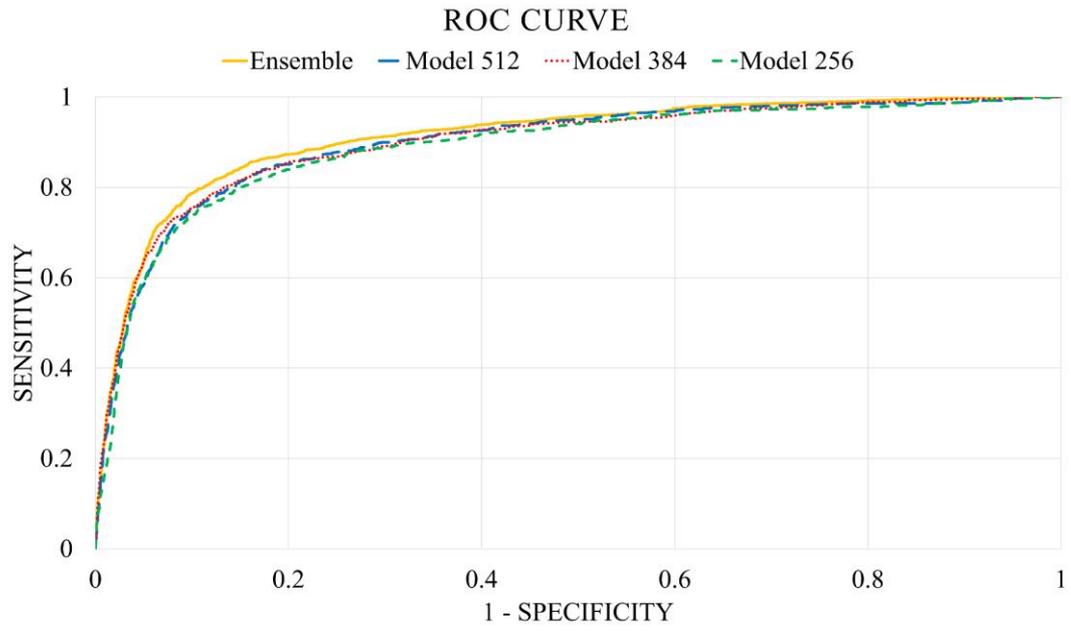

Figure 5. ROC curves for each model.